%% file: iclr2025_conference.tex
\documentclass{article} 
\usepackage{iclr2025_conference,times}

\input{math_commands.tex}

\usepackage{url}
\input{macro}

\title{RMBoost: Reward Model Training With Preference-Conditional Multi-Aspect Synthetic Data Generation}

\author{Jiaming Shen $^\dagger$ \And
  Ran Xu $^\ddagger$ \And
  Yennie Jun $^\dagger$ \And
  Zhen Qin $^\dagger$ \And
  Tianqi Liu $^\dagger$ \AND
  Carl Yang $^\ddagger$\thanks{Work done while being a visiting faculty in Google.} \And
  Yi Liang $^\dagger$ \And
  Simon Baumgartner $^\dagger$ \And
  Michael Bendersky $^\dagger$ \And
  \normalfont{Google$^\dagger$, Emory University$^\ddagger$} \\
  \texttt{\{jmshen, zhenqin, tianqiliu, yiliang, simonba, bemike\}@google.com} \\
  \texttt{\{ran.xu, j.carlyang\}@emory.edu} \\
}

\iclrfinalcopy 
\begin{document}

\maketitle
\begin{abstract}

Reward models (RMs) are crucial for aligning large language models (LLMs) with human preferences.
They are trained using preference datasets where each example consists of one input prompt, two responses, and a preference label.
As curating a high-quality human labeled preference dataset is both time-consuming and expensive, people often rely on existing powerful LLMs for preference label generation.
This can potentially introduce noise and impede RM training.
In this work, we present \ours, a novel synthetic preference data generation paradigm to boost reward model quality.
Unlike traditional methods, which generate two responses before obtaining the preference label, \ours first generates one response and selects a preference label, followed by generating the second more (or less) preferred response conditioned on the pre-selected preference label and the first response.
This approach offers two main advantages.
First, \ours reduces labeling noise since preference pairs are constructed intentionally.
Second, \ours facilitates the creation of more diverse responses by incorporating various quality aspects (e.g., helpfulness, relevance, completeness) into the prompts.
We conduct extensive experiments across three diverse datasets and demonstrate that \ours outperforms other synthetic preference data generation techniques and significantly boosts the performance of four distinct reward models.

\end{abstract}

\input{002-introduction}
\input{003-related_work}

\input{004-preliminaries}

\input{004-method}
\input{005-experiment}
\input{006-conclusion}

\bibliography{iclr2025_conference}
\bibliographystyle{iclr2025_conference}

\appendix
\input{007-appendix}

\end{document}

%% file: math_commands.tex
\usepackage{amsmath,amsfonts,bm}

\def\eqref#1{equation~\ref{#1}}

\def\1{\bm{1}}

\DeclareMathAlphabet{\mathsfit}{\encodingdefault}{\sfdefault}{m}{sl}
\SetMathAlphabet{\mathsfit}{bold}{\encodingdefault}{\sfdefault}{bx}{n}



%% file: macro.tex
\usepackage[utf8]{inputenc} 
\usepackage[T1]{fontenc}    
\usepackage[colorlinks]{hyperref}       
\usepackage{url}            
\usepackage{booktabs}       
\usepackage{amsfonts}       
\usepackage{nicefrac}       
\usepackage{microtype}      
\usepackage{xcolor}         

\definecolor{green}{RGB}{48,128,20}
\usepackage{titlesec}
\usepackage{textcomp,booktabs}

\usepackage{colortbl}
\definecolor{pink}{rgb}{.99,.91,.95}
\usepackage{graphics}
\usepackage{multirow}
\usepackage{pdfpages}
\usepackage{amsmath, amssymb, amsthm, xspace, color}

\usepackage{graphicx} 
\usepackage{float} 
\usepackage{bm}
\usepackage{subfigure}
\usepackage{enumitem}
\usepackage{wrapfig}

\usepackage{xspace}
\newcommand{\ours}{\texttt{RMBoost}\xspace}

\newcommand{\Prb}{\operatorname{Pr}}

\usepackage{listings}
\lstdefinestyle{mystyle}{
  basicstyle=\ttfamily\small,
  frame=single,
  breaklines=true,
  breakindent=1pt,
  backgroundcolor=\color{blue!2}, 
}

%% file: 002-introduction.tex

\section{Introduction}\label{sec:intro}

Large language models (LLMs)~\citep{anil2023palm, Achiam2023GPT4TR, Anil2023GeminiAF, Touvron2023Llama2O} have recently demonstrated unprecedented capabilities in various tasks.
Leveraging a reward model (RM) to align LLMs with human preference (either through reinforcement learning~\citep{Ouyang2022TrainingLM, stiennon2020learning} or via direct optimization over RM offline labeled preference pairs~\citep{Liu2024StatisticalRS,Rafailov2023DirectPO,yuan2024rrhf}) is widely considered as a major breakthrough in modern LLM developments, when traditional supervised fine-tuning (SFT) alone yields suboptimal generation quality~\citep{kirk2024understanding}. 

\begin{figure}[!t]
    \centering
    \vspace{-1.0ex}
    \includegraphics[width=0.98\linewidth]{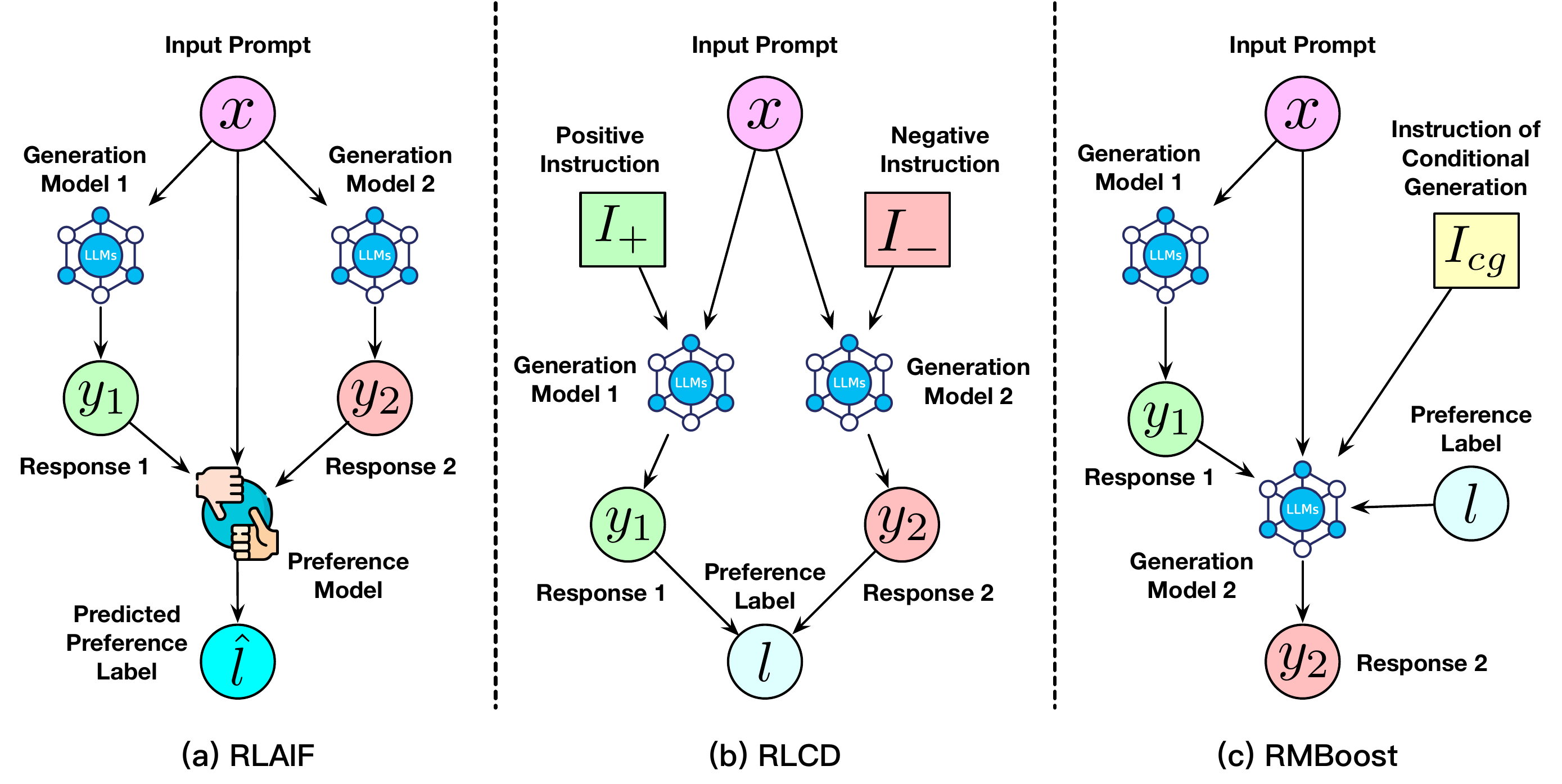}
    \vspace{-1.0ex}
    \caption{The illustration compares \ours with existing paradigms for synthetic preference data generation. The two generation models can correspond to the same model (configured at non-zero temperature) that samples two different responses.
    \textbf{(a)} The RLAIF approach first generates two responses and then leverages an initial preference model (e.g., LLM with few-shot side-by-side demonstrations) to predict the preference label.
    \textbf{(b)} The RLCD method produces two responses using contrasting prompts and assigns the preference label based on the respective prompt.
    \textbf{(c)} Our approach uniquely generates the second response conditioned on the first response and a predetermined preference label.
    }
    \vspace{-2.0ex}
    \label{fig:overview}
\end{figure}

To develop a high-quality reward model, it is necessary to collect a preference dataset comprising triplets of input prompt $x$, dual responses ($y_1$, $y_2$), and a response preference label $l$.
As manually curating such preference dataset at scale is expensive, researchers have investigated automated methods for generating preference labels.
One pioneering work, RLAIF~\citep{Bai2022ConstitutionalAH}, proposes to generate synthetic preference labels by prompting an LLM with few-shot side-by-side demonstrations (see Fig.~\ref{fig:overview}(a)).
Follow up studies~\citep{Lee2023RLAIFSR, Pace2024WestofNSP} extend this idea by first distilling LLM few-shot predictions into an initial RM and then leveraging it to score candidate response pairs for selecting the preferred response.
However, since the few-shot LLM predictor is not flawless, all these methods produce noisier pairwise preferences than human-generated ones.
In an alternative approach, RLCD~\citep{Yang2023RLCDRL} generates two responses with two contrasting prompts and obtains the preference label based on the prompt employed (see Fig.~\ref{fig:overview}(b)).
While this method reduces the preference labeling errors, these contrasting prompts typically focus on a single evaluation aspect (e.g., helpfulness), which restricts the diversity of the responses generated.

In this work, we present \ours, a novel synthetic preference data generation paradigm designed to boost the quality of reward models (see Fig.~\ref{fig:overview}(c)). 
Our key innovations lie in the \emph{progressive} way of generating preference pairs.
Instead of predicting the preference label $l$ for a pair of existing responses ($y_1$, $y_2$), \ours first generates one response $y_1$ and selects a preference label $l$.
Then, \ours generates a second more (or less) preferred response $y_2$, conditioned on $y_1$ and $l$, and guided by predefined evaluation aspects (e.g., helpfulness, relevance, faithfulness, etc).
In other words, \ours explicitly improves (or corrupts) the first response $y_1$ and transforms it into the second response $y_2$, thereby reducing preference label noise.
Meanwhile, \ours leverages multi-aspect prompting to ensure that $y_2$ is sufficiently distinct from $y_1$, which not only provides fine-grained control over the generated text, but also helps to promote the diversity of generated datasets.

Intuitively, \ours improves performance over previous approaches based on two key observations:
(1) existing methods encounter preference prediction errors when the model needs to weigh multiple evaluation aspects~\citep{hong2023on,knox2024models}, and 
(2) LLMs exhibit strong conditional generation capabilities when provided with specific instructions~\citep{Ouyang2022TrainingLM}. 
For instance, instructing an LLM to either corrupt or improve a response (i.e., $y_1$) with respect to one or more aspects typically yields highly effective results.
Furthermore, \ours benefits from a balance between response distribution shift and label noise.
Previous methods, which sample the second response $y_2$ from the same distribution as the LLM at inference time, tend to introduce more noise in the preference label prediction stage.
In contrast, \ours samples $y_2$ from a modified distribution, which leads to a distribution shift but also reduces the preference prediction errors.
In situations where the distribution shift is minor compared to the benefits of fewer preference errors, our method can significantly improve RM training.

We conduct extensive experiments between \ours and other leading synthetic preference data generation methods on three diverse datasets: QA Feedback~\citep{Wu2023FineGrainedHF}, Ultra Feedback~\citep{Cui2023UltraFeedbackBL}, and TLDR summarization~\citep{stiennon2020learning}.
As detailed in $\S$\ref{sec:experiments}, when generating the preference data with PaLM 2-L~\citep{anil2023palm} and GPT-4~\citep{Achiam2023GPT4TR}, \ours substantially outperforms the established baselines in terms of preference prediction accuracy across four different RM backbones.
Besides, our predictive gains can be successfully propagated to the alignment task: the LLM trained with our \ours \emph{consistently} achieves higher win rate over baselines. 
Our analysis further confirms the benefit of \ours for improving the diversity of the responses and mimicing the style of ground-truth preference pairs.

%% file: 003-related_work.tex

\section{Related Work}\label{sec:related_work}

\noindent \textbf{Synthetic Preference Data Generation.} 
As reward models play a vital role in LLM developments~\citep{Ouyang2022TrainingLM,Touvron2023Llama2O} and it is expensive to collect human preference data for RM training, a few works have leveraged LLMs for synthetic preference data generation.
One early attempt, reinforcement learning from AI feedback (RLAIF)~\citep{Bai2022ConstitutionalAH, Lee2023RLAIFSR}, prompts LLMs to rate preference labels for a given response pair.
Later, the West-of-N~\citep{Pace2024WestofNSP} technique proposes to further bootstrap an existing RM by directly selecting the best and worst candidates in a pool of responses to a given query as the preference pairs.
Along another line of work, ALMoST~\citep{Kim2023AligningLL} proposes to query two LLMs of different qualities and assumes the response from a stronger model is preferred over the response from a weaker model.
More recently, RLCD~\citep{Yang2023RLCDRL} improves RLAIF and ALMoST by adopting two contrasting prompts (one positive, one negative) to generate two responses and directly obtaining preference labels based on the prompts used.
These works all demonstrate that LLMs can generate useful preference data for training reward models.
At a high level, \ours is more related to RLCD, as both skip the preference label prediction step.
RLCD achieves this implicitly via the contrasting prompts along one considered aspect (e.g., helpfulness or harmlessness).
Our method, on the other hand, accomplishes this explicitly by feeding the preference labels directly into the prompt and enabling the LLMs to edit one response along multiple aspects.
See Appendix \ref{app:extended_related_work} for a more extended review of related studies.

\smallskip
\noindent \textbf{Attribute-aware Text Generation.}
Our work is also related to aspect/attribute-controlled text generation.
One pioneering work~\citep{logeswaran2018content} shows that we can modify the style of a sentence while preserving its content using a small neural generation model.
Follow up studies~\citep{russo2020control,yu2021attribute} extend this idea to sentiment and topic controlled text generation.
Based on these findings, a more recent study~\citep{Yu2023LargeLM} proposes AttrPrompt, which leverages LLMs to generate synthetic data for classification tasks.
Our method is related to AttrPrompt in the sense that we all aim to increase the diversity of generated text by leveraging multi-aspect controlled generation. 
However, we differ significantly in terms of the targeted aspects, downstream tasks and input formats
(i.e., sentiment/topic for single-sentence classification tasks in AttrPrompt versus helpfulness/relevance for RM training in \ours).

%% file: 004-preliminaries.tex
\section{Preliminaries}\label{sec:related_work}
In this work, we use $\mathcal{X}$ and $\mathcal{Y}$ to denote the space of model input prompts and model output responses.
Furthermore, we denote the language model to be aligned with human preference as $\pi: \mathcal{X} \rightarrow \mathcal{Y}$ and represent the pointwise reward model as $r: \mathcal{X} \times \mathcal{Y} \rightarrow \mathbb{R} $.

\smallskip
\noindent \textbf{Preference Data and Reward Modeling.}
The reward model $r$ is typically trained on a preference dataset $\mathcal{D}_{\text{RM}} = \{(x^i, y^i_1, y^i_2, l^i)\}|_{i=1}^{N}$ where each example consists of an input prompt $x^i$, two responses $y^i_1, y^i_2$, and a preference label $l^i \in \{-1, 1\}$ indicating which response is preferred.
We denote the preferred response as $y^{i}_{+}$ and the less preferred one as $y^{i}_{-}$.

Following the Bradley-Terry~\citep{Bradley1952RANKAO} assumption, we train the reward model by minimizing the following empirical negative log-likelihood loss:
\begin{equation}\label{eq:rm_loss}
\small
L(r_{\theta}, \mathcal{D}_{\text{RM}}) = -\mathbf{E}_{(x, y_{+}, y_{-}) \in \mathcal{D}_{\text{RM}}} [\log(\sigma(r_{\theta}(x, y_{+}) - r_{\theta}(x, y_{-}) ))],
\end{equation}
where $r_{\theta}$ is the reward model parameterized by $\theta$ and $\sigma$ denotes the sigmoid function.

\smallskip
\noindent \textbf{Synthetic Preference Data Generation.}
The pioneering work~\citep{stiennon2020learning} constructs the preference dataset entirely through manual curation, aptly named ``Reinforcement Learning from \underline{Human Feedback}''.
Later studies~\citep{Bai2022ConstitutionalAH, Lee2023RLAIFSR} propose to replace the human feedback with LLM generated preference labels (in other words, \underline{AI Feedback}).
Specifically, they first assume the access to a set of unlabeled prompts $\{x \in \mathcal{X}\}$.
Then, for each input prompt $x^i$, they generate two responses $y^i_{1}, y^i_{2}$ by sampling from either the same LLM twice or from two different LLMs.
After that, another strong off-the-shelf LLM (e.g., GPT4, Gemini Ultra) is leveraged to generate the synthetic preference label $\hat{l^i}$ (which could be different from the real preference label $l^i$).
Finally, by repeating this process for all input prompts, we obtain a synthetic preference dataset $\mathcal{D}_{\text{SYN}} = \{(x^i, y^i_1, y^i_2, \hat{l^i})\}|_{i=1}^{N}$ for training reward models.

%% file: 004-method.tex
\section{The RMBoost Framework}\label{sec:method}

We first describe our method, \ours, a novel method for boosting reward model training with preference-conditional multi-aspect synthetic data generation.
Then, we present its high-level intuitions, followed by a more theoretical analysis.  

\subsection{Method Description}\label{subsec:method_description}
\begin{figure}[!t]
    \centering
    \vspace{-0.3ex}
    \includegraphics[width=0.99\linewidth]{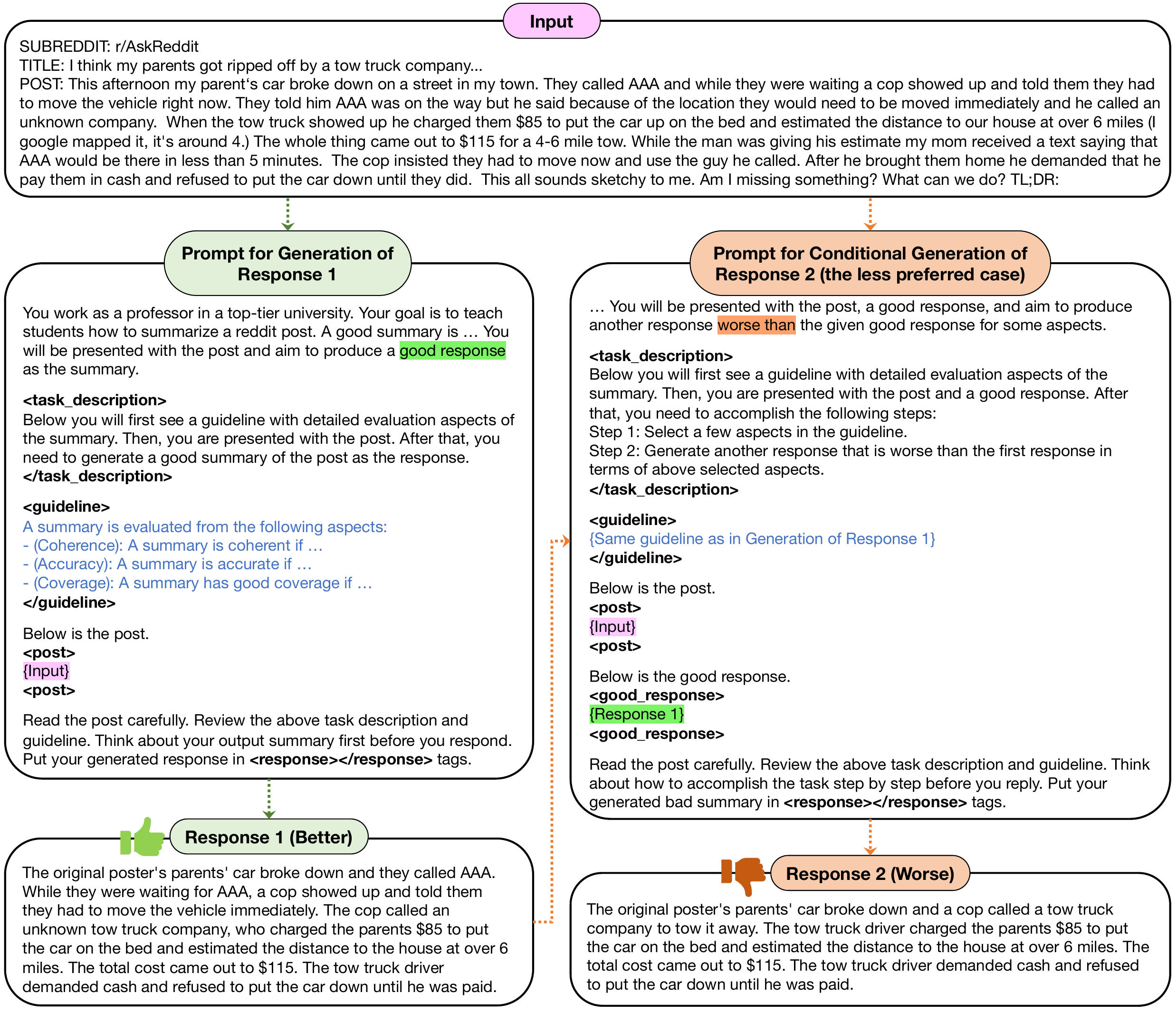}
    \vspace{-1.0ex}
    \caption{Overview of our \ours framework with prompt gists. See $\S$\ref{app:prompts} for the complete prompts.
    }
    \vspace{-1.0ex}
    \label{fig:prompt_fig}
\end{figure}

\ours begins with a collection of unlabeled input prompts $\{x \in \mathcal{X}\}$, similar to  RLAIF or RLCD, and utilizes an off-the-shelf LLM $\pi$ for generating responses. 
For each input prompt $x^i$, our method initially samples one response $y^i_1$ from $\pi$. 
Subsequently, a preference label $l^i$ (designated as ``more preferred'' or ``less preferred'') is predetermined. 
The LLM is then prompted to generate a second response $y^i_2$, this time conditioned on the first response $y^i_1$, the preference label $l^i$, and a specific instruction for \underline{c}onditional \underline{g}eneration $I_{cg}$. 
This instruction outlines all the relevant aspects for evaluating the response, such as helpfulness, relevance, and coherence, and guides the LLMs to adjust the first response according to these criteria. 
Then, we form the preference pair $(y^i_{+}, y^i_{-})$ to be $(y^i_{1}, y^i_{2})$ if the preference label $l^i$ is designated as ``less preferred'', or $(y^i_{2}, y^i_{1})$ if $l^i$ is ``more preferred''.
At last, we can optionally use the standard RLAIF method to check the quality of constructed preference pairs and filter those low quality ones (See Appendix \ref{app:filtering_step_details} for more discussions). 

Figure~\ref{fig:prompt_fig} shows one example where we select the preference label $l $ as ``less preferred'' and intentionally let the LLM corrupt the first tl;dr summary to become worse along the ``coverage'' and ``accuracy'' aspects.
After constructing the synthetic preference data, \ours follows the standard practice to train the reward model.
The learned reward model can then be employed for LLM alignment.

\subsection{Intuitions for RMBoost}
Intuitively, \ours surpasses existing methods based on two key observations: (1) many preference prediction errors in previous methods occur when the model must consider multiple evaluation aspects simultaneously, and (2) the LLM demonstrates robust directional generation capabilities when given explicit instructions. Specifically, when instructed to modify a response (i.e., $y_1$) with respect to one or more aspects, the LLM typically excels.

Another perspective on the effectiveness of \ours is its ability to balance between response distribution shifts and label noise. Traditional methods sample the second response $y_2$ from the same distribution $\Pr(y | x)$ as the LLM's inference time distribution.
However, they need another step to prompt the LLM for preference prediction, which often introduces many labeling noise, especially when comparing two responses requires thorough considerations over multiple evaluation aspects (i.e., our first key observation).
In contrast, \ours samples $y_2$ from a modified distribution $\Pr(y | x, y_1, l, I_{cg})$, resulting in a distribution shift but experiencing fewer preference prediction errors (due to our second key observation). 
If this distribution shift is relatively minor compared to the reduction in preference prediction errors, our method can significantly enhance RM training.
We present this analysis more formally below.

\subsection{Analysis of RMBoost and Comparisons with Previous Approaches}\label{subsec:math_analysis}
Reward model training on a clean human-labeled preference dataset can be viewed as minimizing an empirical version of the following risk function:
\begin{equation}\label{eq:clean_risk}
\small
\mathbf{E}_{(x, y_1, y_2, l) \sim \Prb^{*}} [L(x, y_1, y_2, l ; \theta)],
\end{equation}
where $l$ is the \emph{ground truth} preference label, $L$ is the loss function, $\theta$ is the RM parameter set, and $\Pr^{*}$ denotes the true (human-labeled) data distribution, defined as follows\footnote{\footnotesize Without the loss of generality, we assume two responses are sampled from the same LLM independently. If they come from two different LLMs, we can replace the second $\Prb(y_2|x)$ with another distribution and the below derivations should still hold.}:
\begin{equation}
\small
\Prb^{*} (x, y_1, y_2, l) = \Prb(x)\Prb(y_1|x)\Prb(y_2|x)\Prb^{*}(l|x, y_1, y_2).
\end{equation}
This joint distribution reflects the common human and AI preference data collection processes.

\smallskip
\noindent \textbf{RLAIF.} 
When the preference labels are machine generated, we use $\tilde{l}$ to denote the noisy preference label and follow previous literature~\citep{Sukhbaatar2014TrainingCN} to assume a noise corruption distribution $Q(\tilde{l}|l)$. 
Then, we can define the risk on this ``noisy'' preference dataset as follows:
\begin{equation}\label{eq:noisy_risk}
\small
\mathbf{E}_{(x, y_1, y_2, \tilde{l}) \sim \Prb^{\text{RLAIF}}} [L(x, y_1, y_2, \tilde{l} ; \theta)],
\end{equation}
where the noisy data distribution (for RLAIF approaches) is:
\begin{equation}\label{eq:noise_prob}
\small
\Prb^{\text{RLAIF}} (x, y_1, y_2, \tilde{l}) = \sum_{l} \Prb(x)\Prb(y_1|x)\Prb(y_2|x)\Prb^{*}(l|x, y_1, y_2)Q(\tilde{l}|l).
\end{equation}
By plugging in Eq.~\ref{eq:noise_prob} into the above Eq.~\ref{eq:noisy_risk}, we have:
\small
\begin{align}
\mathbf{E}_{(x, y_1, y_2, \tilde{l}) \sim \Prb^{\text{RLAIF}}} [L(x, y_1, y_2, \tilde{l} ; \theta)] & = 
\mathbf{E}_{(x, y_1, y_2, \tilde{l}) \sim \Prb^{*}} \left[ \beta^{\text{RLAIF}}(x, y_1, y_2, \tilde{l})  L(x, y_1, y_2, \tilde{l} ; \theta) \right] \label{eq:noisy_risk_rewrite} \\
\beta^{\text{RLAIF}}(x, y_1, y_2, \tilde{l}) &= \frac{\Prb^{\text{RLAIF}} (x, y_1, y_2, \tilde{l})}{\Prb^{*} (x, y_1, y_2, \tilde{l})}.\label{eq:noisy_risk_beta}
\end{align}
\normalsize

The left hand side of Eq.~\ref{eq:noisy_risk_rewrite} is the raw training objective. 
The right hand side of Eq.~\ref{eq:noisy_risk_rewrite} is essentially a sample re-weighted version of above Eq.~\ref{eq:clean_risk}. 
In other words, when we consider the preference label noise, we are optimizing a biased risk function on clean data. 
Furthermore, the closer the re-weighting factor $\beta^{\text{RLAIF}}(x, y_1, y_2, \tilde{l})$ is to 1, the less bias we have on the training objective of existing methods.

Let dive into the re-weighting factor by combining Eq.~\ref{eq:noise_prob} with Eq.~\ref{eq:noisy_risk_beta}, we have:
\begin{equation}\label{eq:noisy_risk_beta_expand}
\small
\beta^{\text{RLAIF}}(x, y_1, y_2, \tilde{l}) = \frac{\sum_{l} \Prb^{*}(l|x, y_1, y_2)Q(\tilde{l}|l) }{ \Prb^{*}(\tilde{l}|x, y_1, y_2) } = Q(\tilde{l} | \tilde{l}) + \frac{\sum_{l \neq \tilde{l}} \Prb^{*}(l|x, y_1, y_2)Q(\tilde{l}|l)  }{\Prb^{*}(\tilde{l}|x, y_1, y_2)}.
\end{equation}
If we consider a simple binary preference prediction setting where either $y_1$ or $y_2$ is preferred without tie. 
Given a noisy example where the observed label $\tilde{l}$ is different from its true label $l$, the second term in the right-hand side of Eq.~\ref{eq:noisy_risk_beta_expand} will explode to a very large number. 
This is because its denominator $\Prb^{*}(\tilde{l}|x, y_1, y_2)$, the ground truth probability of getting a wrong prediction label $\tilde{l}$, is close to 0, while the nominator $\Prb^{*}(l|x, y_1, y_2)$ is close to 1. 
Consequently, $\beta^{\text{RLAIF}}(x, y_1, y_2, \tilde{l})$ is far away from 1 and we are optimizing a very biased version of the true risk function.

\smallskip
\noindent \textbf{RLCD.}
The synthetic preference pairs generated by RLCD follow the below distribution\footnote{\small We assume the first response is more preferred. Due to symmetries, below derivations hold for the case where the second response is more preferred.}:
\begin{equation}\label{eq:rlcd_distribution}
\small
\Prb^{\text{RLCD}} (x, y_1, y_2, l) = \Prb(x)\Prb(l)\Prb(y_1|x, I_{+})\Prb(y_2|x, I_{-}),
\end{equation}
where $I_{+}$ ($I_{-}$) denotes the prompt for generating the positive (negative) response, respectively.
Following the same derivations above, we will have:
\begin{equation}\label{eq:rlcd_beta}
\small
\beta^{\text{RLCD}}(x, y_1, y_2, l) = \frac{\Prb(l)\Prb(y_1|x, I_{+})\Prb(y_2|x, I_{-})}{\Prb(y_1|x)\Prb(y_2|x)\Prb^{*}(l|x, y_1, y_2)} ,
\end{equation}
Here, we first notice that $\Prb(l)$ and $\Prb^{*}(l|x, y_1, y_2)$ are essentially preference label frequency distribution and they won't matter if we adopt the example flipping strategy\footnote{\footnotesize The example flipping strategy means we can flip $y_1$, $y_2$ along with their corresponding preference label and construct a ``new'' reward model training example, which effectively makes both $\Prb(l)$ and $\Prb^{*}(l|x, y_1, y_2)$ to be uniform distribution and cancel out.}.
The remaining terms indicate the distribution shift between real responses and synthetically generated responses.

\smallskip
\noindent \textbf{RMBoost.}
In our synthetic data generation method, we directly sample the true preference label $l$ and generate the second response $y_2$. Therefore, the distribution of our synthetic data is:
\begin{equation}\label{eq:our_prob}
\small
\Prb^{\text{RMBoost}} (x, y_1, y_2, l) = \Prb(x)\Prb(y_1|x)\Prb(l)\Prb(y_2| x, y_1, l, I_{cg}),
\end{equation}
where $I_{cg}$ is a fixed conditional generation instruction.
We then follow the same derivation of Eq.~\ref{eq:noisy_risk_rewrite} and obtain our re-weighting factor as follows:
\begin{equation}\label{eq:our_risk_beta_expand}
\small
\beta^{\text{RMBoost}}(x, y_1, y_2, l) = \frac{\Prb(l)\Prb(y_2| x, y_1, l, I_{cg})}{\Prb^{*}(l|x, y_1, y_2)\Prb(y_2|x)}.
\end{equation}
Similar to the above RLCD derivation, we can cancel out the terms $\Prb(l)$ and $\Prb^{*}(l|x, y_1, y_2)$.
The remaining terms $\Prb(y_2| x, y_1, l, I_{cg})$, $\Prb(y_2|x)$ correspond to the second response distributions.
If these two distributions are close to each other, we will have a $\beta^{\text{RMBoost}}(x, y_1, y_2, l)$ closer to 1, which enables us to optimize a less biased version of the true risk function.
In the below experiment $\S$\ref{subsec:analysis_of_synthetic_data}, we empirically show that \ours indeed produces a distribution of re-weighting factor $\beta$ closer to 1 and facilitates the reward model training with a less biased objective.

%% file: 005-experiment.tex

\section{Experiments}\label{sec:experiments}

\subsection{Experiment Settings}\label{subsec:experiment_settings}

\textbf{Datasets and Tasks.}
We analyze and evaluate all compared methods on three diverse sets of datasets.
(1) \textbf{QA Feedback}~\citep{Wu2023FineGrainedHF} is a long-form QA dataset where the model inputs a question, a set of Wikipedia passages and outputs a long-form response to answer the given question. 
The raw dataset contains both supervised finetune (SFT) data and reward model (RM) training data.  
For SFT data, we re-use its original train/val splits.
For RM data, we construct preference pairs by considering all pairwise responses in the raw data and split its original validation set into new val/test sets.
(2) \textbf{Ultra Feedback}~\citep{Cui2023UltraFeedbackBL} is a large-scale, diversified preference dataset built for general LLM alignment research. Each example in this benchmark includes one prompt and 4 responses associated with their quality scores. 
As the original benchmark does not contain SFT data and its preference data have no standard train/val/test splits, we create our own as follows.
For each example, we first select the response with the best overall score as the candidate preferred response, and the one with the least overall score as the candidate not preferred response. 
Then, we check if the score of the candidate preferred response is larger than the score of the candidate not preferred response by at least 1.5. 
If yes, we construct a preference pair. Otherwise, we place this candidate preferred response along with its prompt into the SFT data.
(3) \textbf{TLDR Summarization}~\citep{stiennon2020learning} consists of Reddit posts along with their human written summaries (for SFT) and pairs of machine-generated summaries rated by human labels (for RM training). 
We use the existing splits in the original dataset.

\smallskip
\noindent \textbf{Compared Methods.}
We compare our \ours method with the following synthetic preference data generation baselines: 
(1) \textbf{RLAIF}~\citep{Bai2022ConstitutionalAH}, which uses LLMs to rate preference labels;
(2) \textbf{West-of-N}~\citep{Pace2024WestofNSP}, which directly selects the best and worst candidates in a pool of responses to a given query;
(3) \textbf{RLCD}~\citep{Yang2023RLCDRL}, which leverages two contrasting prompts to generate two responses and returns the response associated with the positive prompt as the preference response. 
For each compared method as well as \ours, we report their associated prompt templates in Appendix~\ref{app:prompts}.

\smallskip
\noindent \textbf{Backbone LLMs.}
For synthetic preference data generation, we use PaLM 2-L~\citep{anil2023palm} in our main experiments.
We also evaluate \ours with GPT-3.5~\citep{chatgpt} and GPT-4~\citep{Achiam2023GPT4TR} in $\S$\ref{subsec:GPT_main_results}.
For RM training (on both real and synthetic preference data as well as their mixture), we investigate four backbone models: Gemini-Nano-1, Gemini-Nano-2~\citep{Anil2023GeminiAF}, PaLM 2-XXS~\citep{anil2023palm}, and Gemma 2B~\citep{Mesnard2024GemmaOM}.

\smallskip
\noindent \textbf{Evaluation Protocols.}
We first train the model on each dataset's SFT data and select the checkpoint with the highest performance metric on the SFT validation set.
Then, we initialize the reward model with the above selected SFT checkpoint and continue fine-tune it on the reward model training set.
We use each dataset's RM validation set to select hyper-parameters and report the preference prediction accuracy of each fine-tuned RM on the RM test set.
Appendix~\ref{app:implementation_details} includes more details.

\begin{table*}[!t]
    \centering
    \renewcommand\arraystretch{1.0}
 \caption{Statistics of datasets.}
\resizebox{0.99\linewidth}{!}{
\begin{tabular}{ll|cc|ccc}
\toprule
    \bf Dataset & \bf Task & \bf \# SFT Train & \# \bf SFT Val & \bf \# RM Train &\bf  \# RM Val &\bf  \# RM Test  \\\midrule
    QA Feedback~\citep{Wu2023FineGrainedHF} & Question Answering       & 1,000 & 500 & 14,982 & 1,344  & 1,272 \\
    Ultra Feedback~\citep{Cui2023UltraFeedbackBL}  & General LLM Alignment      & 13,920 & 8,477 & 33,897 &  4,238  & 4,239 \\
    TLDR Summarization~\citep{stiennon2020learning}  & Summarization   & 116,722 & 6,447 & 92,534 & 83,797  & 83,629  \\
    \bottomrule
\end{tabular}
}
\label{tab:dataset}
\end{table*}

\begin{table*}[!tbp]
  \centering
  \caption{Overall experiment results with PaLM 2-L as the preference data generation model. We train each reward model backbone on preference data generated by each method. Specifically, ``Real'' (``Syn'') indicates that RM is trained only on the real (synthetic) preference data and ``Real + Syn'' means the RM is fine-tuned on a mixture of real and synthetic preference data. All numbers are the preference prediction accuracy of a fine-tuned RM on each dataset's test set.}
  \label{tab:PaLM2L_main_exp_results}
  \resizebox{1.0\linewidth}{!}{
  \begin{tabular}{ll|ccc|ccc|ccc}
  \toprule
  \multirow{2}{*}{\textbf{Backbone}} & \multirow{2}{*}{\textbf{Methods}} & \multicolumn{3}{c}{\bfseries QA Feedback} & \multicolumn{3}{c}{\bfseries Ultra Feedback} & \multicolumn{3}{c}{\bfseries TLDR Summarization} \\
  \cmidrule(lr){3-5} \cmidrule(lr){6-8} \cmidrule(lr){9-11}
   &  & Real & Syn & Real + Syn & Real & Syn & Real + Syn & Real & Syn & Real + Syn \\
  \midrule
  \multirow{4}{*}{Gemini-Nano-1}
  & RLAIF           & \multirow{4.3}{*}{63.67} & 56.25 & 61.33 & \multirow{4.3}{*}{79.69} & 62.89 & 76.17 & \multirow{4.3}{*}{74.61} & 68.75 & 73.83 \\
  & West-of-$N$     &  & 59.38 & 60.94 &  & 73.05 & 88.28 &  & 70.31 & 75.00 \\
  & RLCD            &  & 60.16 & 64.06 &  & 73.44 & 88.67 &  & 70.70 & 76.56 \\
  & \ours           &  & \bf 61.33 & \bf 67.97 &  & \bf 74.61 & \bf 90.23 &  & \bf 71.48 & \bf 77.34 \\
  \midrule
  \multirow{4}{*}{Gemini-Nano-2} 
  & RLAIF           & \multirow{4.3}{*}{67.58} & 59.38 & 65.23 & \multirow{4.3}{*}{92.97} & 75.00 & 89.06 & \multirow{4.3}{*}{80.47} & 71.48 & 79.69 \\
  & West-of-$N$     &  & 60.55 & 65.63 &  & 76.56 & 90.63 &  & 71.88 & 80.08 \\
  & RLCD            &  & \bf 61.33 & 66.80 &  & \bf 79.69 & 93.75 &  & \bf 73.83 & 81.26 \\
  & \ours           &  & 60.55 & \bf 68.36 &  & 79.30 & \bf 94.14 &  & 73.44 & \bf 82.81 \\
  \midrule
  \multirow{4}{*}{PaLM 2-XXS} 
  & RLAIF           & \multirow{4.3}{*}{70.31} & 57.81 & 64.06 & \multirow{4.3}{*}{90.63} & 75.26 & 89.56 & \multirow{4.3}{*}{71.48} & 63.28 & 70.31 \\
  & West-of-$N$     &  & 59.38 & 67.17 &  & 76.56 & 92.97 &  & 64.45 & 71.09 \\
  & RLCD            &  & 60.94 & 70.31 &  & \bf 78.13 & \bf 93.75 &  & \bf 67.67 & 72.27 \\
  & \ours           &  & \bf 64.06 & \bf 75.00 &  & \bf 78.13 & \bf 93.75 &  & 66.80 & \bf 72.66 \\
  \midrule
  \multirow{4}{*}{Gemma 2B} 
  & RLAIF           & \multirow{4.3}{*}{60.05} & 51.17 & 56.64 & \multirow{4.3}{*}{86.28} & 60.94 & 82.81 & \multirow{4.3}{*}{70.69} & 62.50 & 67.67 \\
  & West-of-$N$     &  & 54.30 & 57.03 &  & 65.63 & 85.94 &  & 63.28 & 70.70 \\
  & RLCD            &  & 55.86 &  60.55 &  & 68.65 & 87.50 &  & 64.45 & 71.48 \\
  & \ours           &  & \bf 56.92 & \bf 61.47 &  & \bf 68.71 & \bf 87.88 &  & \bf 65.65 & \bf 71.53 \\
  \bottomrule
  \end{tabular}
  }
\end{table*}

\vspace{-0.5em}
\subsection{Main Experiment Results}\label{subsec:PaLM2L_main_results}
\vspace{-0.5em}
In our main experiments, we first use different synthetic data generation methods to generate preference datasets.
Then, we train various backbone reward models on three dataset settings: (1) \textbf{Real}: only human rated dataset, (2) \textbf{Syn}: only synthetically generated dataset, and (3) \textbf{Real + Syn}: mixture of both human rated and synthetically generated datasets.
Results are shown in Table~\ref{tab:PaLM2L_main_exp_results}, from which we have the following findings.
We observe that mixing synthetic data with human labeled ones in general can help the reward model training.
Among all the methods, \ours generally outperforms other baseline methods in both synthetic and mixtures variants, indicating the high quality of generated synthetic preference data.
Besides, when mixing output synthetic data with the real data, \ours gives the most performance boost, which highlights that \ours effectively complements the ground-truth data to strengthen reward modeling.

\begin{table}[!t]
\begin{minipage}{.38\linewidth}
\caption{Ablations of \ours with Gemini-Nano-1 as the backbone RM. 
}
\label{table:rmboost_ablation_studies}
\resizebox{\columnwidth}{!}{
\begin{tabular}{l|ccc}
\toprule
\multirow{2}{*}{\textbf{Method}} & \bf QA & \bf Ultra & \bf TLDR  \\
                                & \bf Feedback & \bf Feedback & \bf Summarization  \\
\midrule
\ours                       & 67.97 & \bf 90.23 & 77.34 \\
$~~$No-Aspect               & 63.67 & 88.28 & 75.78 \\
$~~$No-Filtering            & 67.41 & 89.06 & 77.34 \\
$~~+$SFT-Response        & \bf 68.75 & 89.45 & \bf 79.30 \\
\bottomrule
\end{tabular}
}

\end{minipage}
\hfill
\begin{minipage}{.58\linewidth}
\caption{\ours with GPT-3.5 and GPT-4 as synthetic data generation (DGen) models.
}
\label{table:GPT_results}
\resizebox{\columnwidth}{!}{
  \begin{tabular}{ll|ccc|ccc}
  \toprule
  \multirow{2}{*}{\textbf{RM Backbone}} & \multirow{2}{*}{\textbf{DGen Model}} & \multicolumn{3}{c}{\bfseries QA Feedback} & \multicolumn{3}{c}{\bfseries TLDR Summarization} \\
    & & Real & Syn & Real + Syn & Real & Syn & Real + Syn  \\
  \midrule
  \multirow{2}{*}{Gemini-Nano-1}
    & GPT-3.5   & \bf 63.67 & 58.59 & 63.28 & 74.61 & 71.88 & \bf 77.73  \\
    & GPT-4     & 63.67 & 60.16 & \bf 66.02 & 74.61 & 73.05 & \bf 78.52  \\
  \midrule
  \multirow{2}{*}{Gemini-Nano-2} 
    & GPT-3.5   & 67.58 & 60.94 & \bf 67.91 & 80.47 & 72.27 & \bf 81.97 \\
    & GPT-4     & 67.58 & 64.06 & \bf 68.75 & 80.47 & 73.83 & \bf 83.26 \\
  \bottomrule
  \end{tabular}

}

\end{minipage}
\vspace{-0.5em}
\end{table}

\subsection{Ablations of RMBoost}\label{subsec:ablations}
We continue to evaluate a few variants of \ours and study how different design choices affect its performance.
First, we compare \ours with its \textbf{``No-Aspect''} version where we intentionally remove the detailed aspect definitions in response generation instruction and replace them with general words like ``good/bad response''.
Second, we test a \textbf{``No-Filtering''} variant which skips the post generation quality check step (see $\S$\ref{subsec:method_description}) and directly use all generated preference pairs to train reward models.
We run these ablation studies with Gemini-Nano-1 as the backbone model and report the results in Table~\ref{table:rmboost_ablation_studies}.
We can see that both variants perform worse than the full \ours by varying degrees.
Specifically, we observe that removing the multiple aspect definitions will significantly hurt the quality of generated data (and thus the RM quality).
Meanwhile, skipping the post-generation quality check step has smaller (though negative) effects can could be considered as an option when the computation resources are limited (for LLM bulk inference).

Additionally, we test a variant of \ours that directly leverages the SFT response as the ``preferred'' response and generates a less preferred one as its counterpart.
We denote this variant as ``\textbf{+SFT-Response}''.
From Table~\ref{table:rmboost_ablation_studies}, we can see that this variant can outperform the vanilla \ours on two out of three datasets.
Although this improvement is somewhat expected as additional high-quality signals from the SFT data are introduced, we think this could be an interesting variant for two reasons.
First, it allows us to skip the first response generation step and thus saves the computes.
Second, it enables people to construct a preference dataset from an SFT dataset by utilizing the conditional generation module in \ours.

\subsection{Experiments with More Synthetic Data Generation Models}\label{subsec:GPT_main_results}
We showcase the versatility of \ours by employing GPT-3.5~\citep{chatgpt} and GPT-4~\citep{Achiam2023GPT4TR} as data generation models. Specifically, we utilize Gemini-Nano-1 and Gemini-Nano-2 as the backbone reward models, training them on the datasets generated above. Table~\ref{table:GPT_results} presents the results on two datasets. From the results, we observe that employing a more powerful LLM as the backbone enhances the quality of synthetic preference data, aligning with the intuition that larger models excel in instruction comprehension. 
Furthermore, we note that \ours serves as a versatile data generation pipeline, adaptable to different generation models, thus producing high-quality data to enhance reward model performance across diverse datasets.

\subsection{Analysis of Generated Synthetic Data}\label{subsec:analysis_of_synthetic_data}
Below, we analyze the quality of our synthetically generated datasets from two aspects.
First, as discussed in $\S$\ref{subsec:math_analysis}, RM training on the synthetic preference data is equivalent to minimizing a biased risk function.
The bias is determined by a per-sample re-weighting factor $\beta$.
Therefore, we first analyze the distributions of this factor $\beta$ by evaluating \ours and RLCD (the best performing baseline) with Gemini-Nano-1 as the backebone RM.
Results are plotted in Figure~\ref{fig:beta_hist_winrate}(a). 
We can see that \ours overall produces synthetic preference datasets that are more similar to the real datasets (the median $\beta$ is closer to 1).
Furthermore, the responses generated from \ours have a larger range of $\beta$ value, which indicates that they are more diverse.

Second, we analyze the length of generated preference pairs.
Specifically, we compute the length ratio of a preferred (win) response over a less preferred (lose) response.
Figure~\ref{fig:beta_hist_winrate}(b) plots a histogram of these length ratios on the QA Feedback dataset.
We observe that both our method and RLCD generally prefer longer responses than shorter ones (as the modes of both methods are larger than 1).
In addition, we can see that the length ratio distribution of \ours generations is closer to the real data, which partially explains the success of our approach.

\subsection{Cost and Efficiency of RMBoost}\label{subsec:cost_rmboost}
To label each synthetic preference example, \ours will call the LLM twice for generating two responses, similar to RLCD.
However, both the input prompt (which includes the first generated response) and the output response of \ours is longer than RLCD, which leads to longer decoding time and larger cost.
Empirically, we observe that the total inference cost of \ours, measured in the number of tokens, is about 30\% larger than RLCD.
This overhead can be potentially reduced when we leverage a pre-existing response (e.g., the SFT response).

\subsection{Measuring RM Quality for LLM Alignment}\label{subsec:rm_for_rlhf}
\begin{figure}[!t]
    \centering
    \vspace{-1.0ex}
    \includegraphics[width=1.0\linewidth]{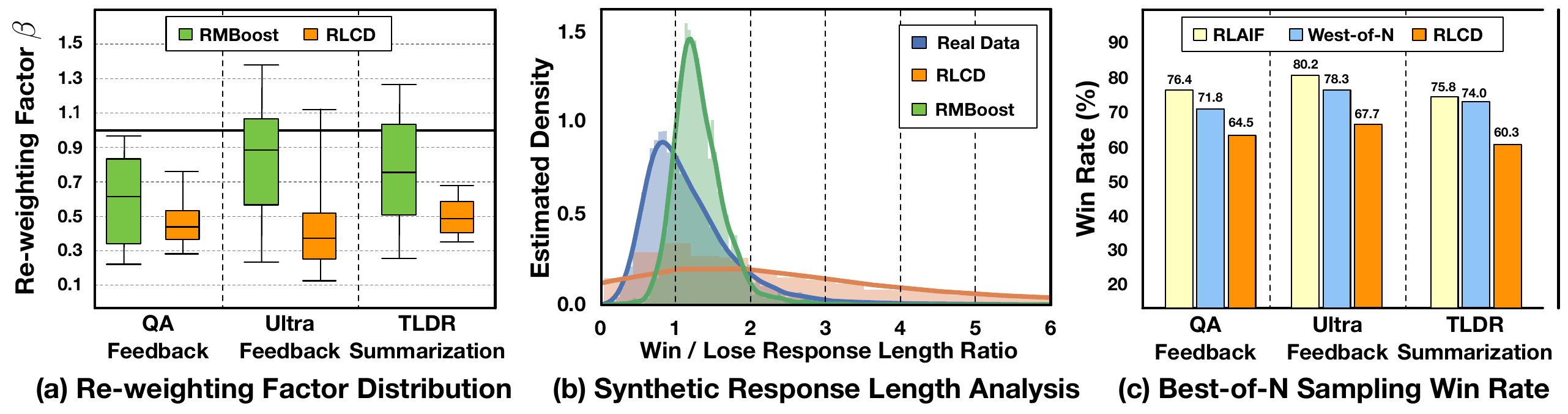}
    \vspace{-1.0ex}
    \caption{
    \textbf{(a)} Distributions of response re-weighting factor $\beta$.
    \textbf{(b)} Histograms of win and lose response length ratio.
    \textbf{(c)} The win rates of \ours over compared baselines for aligning LLMs.
    }
    \vspace{-1.0ex}
    \label{fig:beta_hist_winrate}
\end{figure}

We continue to evaluate the quality of these RMs for aligning LLMs.
For each dataset, we employ the \emph{best-of-N} sampling by first leveraging the SFT checkpoint to generate 9 responses, and then selecting the one with the highest reward score from each trained RM.
Then, we train a large Gemini-Pro model on each dataset's original RM training data and utilize it as the side-by-side auto-rater.
Finally, we compare the responses selected by \ours with those selected by each individual RM using this auto-rater and report the results in Figure~\ref{fig:beta_hist_winrate}(c).
We observe that the RM trained on \ours can generally select better responses than RMs trained on other synthetic data generation methods, and such gains can be further propagated to downstream alignment tasks. 
This demonstrates that \ours enables developer to obtain a better reward model for aligning LLMs.

%% file: 006-conclusion.tex

\section{Limitations and Impact \& Ethics Statement}\label{sec:limitation_and_impact}
\vspace{-0.5em}
One limitation of \ours is that it requires many computational resources for LLM bulk inferences.
In addition, \ours assumes that prompt engineers are fully aware of all evaluation criteria when crafting multi-aspect generation instructions.
Finally, like many other synthetic data generation methods, \ours could potentially generate plausible but inaccurate information.
Overall, \ours has a positive social impact by enabling developers to improve reward models, thereby better aligning LLMs with human values. 
However, there is a risk that \ours could be misused to introduce biases or harmful contents into synthetic datasets intentionally.

\section{Conclusions and Future Work}\label{sec:conclusion_and_future_work}
\vspace{-0.5em}
This study introduces \ours, an innovative approach to generating synthetic preference data for training RMs.
By systematically generating responses conditioned on pre-selected preference labels, \ours effectively minimizes the label noise, a common limitation in previous methods. 
Our experiments validate that \ours enhances the accuracy of RMs across various datasets, establishing a new approach for synthetic preference data generation.
Looking ahead, future research could further optimize \ours by exploring additional dimensions of response generation and refinement. 
Additionally, expanding \ours to domains with multi-model data inputs could be interesting direction.
Ultimately, enhancing the robustness and scalability of \ours will be crucial for its adoption in broader LLM applications, making this an exciting direction for subsequent advancements in the field.

%% file: 007-appendix.tex
\UseRawInputEncoding

\section{More Related Work}\label{app:extended_related_work}

\noindent \textbf{Synthetic Preference Data Generation.} 
LLMs have proven effective for serving as training data generators across diverse applications, but most works emphasize on task-specific finetuning~\citep{meng2022generating,Shao2023SyntheticPG,ye2022zerogen,yu2023regen}, instruction finetuning~\citep{Wang2022SelfInstructAL,honovich2023unnatural,peng2023instruction,xu2023wizardlm,li2024synthetic,Liu2024BestPA} and alignment~\citep{dong2024self,yuan2024selfrewarding}.
As the reward model (RM) plays a vital role in LLM developments~\citep{Ouyang2022TrainingLM,Touvron2023Llama2O} and it is expensive to collect human preference data for RM training, a few recent works have leveraged LLMs for synthetic preference data generation.
One early attempt, known as reinforcement learning from AI feedback (RLAIF)~\citep{Bai2022ConstitutionalAH, Lee2023RLAIFSR}, uses LLMs with few-shot demonstrations to rate preference labels for a given response pair.
Later, the West-of-N~\citep{Pace2024WestofNSP} technique was developed to further bootstrap an existing RM by directly selecting the best and worst candidates in a pool of responses to a given query as the preference pairs.
Along another line of work, ALMoST~\citep{Kim2023AligningLL} proposes to query two LLMs of different qualities and assumes the response from a stronger model (e.g., GPT-4) is preferred over the response from a weaker model (e.g., LLaMA-7B).
More recently, RLCD~\citep{Yang2023RLCDRL} improves RLAIF and ALMoST based on the context distillation idea~\citep{sun2023principle}.
Specifically, RLCD adopts two contrasting prompts (one positive, one negative) to generate two responses and directly obtains preference labels based on the prompts used.
These works all demonstrate that LLMs can generate useful preference data for training reward models.
At a high level, \ours is more related to RLCD, as both skip the preference label prediction step.
RLCD achieves this implicitly via the contrasting prompts along one considered aspect (e.g., helpfulness or harmlessness).
Our method, on the other hand, accomplishes this explicitly by feeding the preference labels directly into the prompt and enabling the LLMs to edit one response along multiple aspects.

\smallskip
\noindent \textbf{Attribute-aware Text Generation.}
Our study is also related to aspect/attribute-controlled text generation.
One pioneering work~\citep{logeswaran2018content} shows that we can modify the style of a sentence while preserving its content using a small neural generation model.
Follow up studies~\citep{russo2020control,yu2021attribute} extend this idea to sentiment and topic controlled text generation.
Based on these findings, a more recent study~\citep{Yu2023LargeLM} proposes AttrPrompt, which leverages LLMs to generate synthetic data for classification tasks.
Our method is related to AttrPrompt in the sense that we all aim to increase the diversity of generated text by leveraging multi-aspect controlled generation. 
However, we differ significantly in terms of the targeted aspects, downstream tasks and input formats
(i.e., sentiment/topic for single-sentence classification tasks in AttrPrompt versus helpfulness/relevance for RM training in \ours).

\section{Optional Preference Pair Filtering Step Details}\label{app:filtering_step_details}
After constructing the preference pairs via \ours, we can optionally perform an additional quality check step.
This step aims to reduce potential sampling errors from the conditional generation phase and to filter those low quality pairs.
It is fully automated --- we use the standard RLAIF method to verify the preference order of generated pairs, retaining only those that match the order established by \ours.
In our experiments, we found the filtering rate is approximately 25\% for \ours. 
We also added the same filtering step to RLCD and found that the filtering rate for RLCD is significantly lower (less than 5\%), likely because RLCD modifies the response by only one aspect at a time. 

\section{Implementation Details}\label{app:implementation_details}
\label{app:imp_details}

For Gemini-Nano-1, Gemini-Nano-2, and PaLM 2-XXS are the RM backbones, we do parameter swamping on the learning rate in [3e-6, 1e-6, 3e-5, 1e-5, 3e-4, 1e-4], batch size in [8, 16, 32].
On both QA Feedback and Ultra Feedback datasets, the final selected hyper-parameters for Gemini-Nano-1 are learning rate = 1e-4, batch size = 16. 
For Gemini-Nano-2, learning rate = 3e-6, batch size = 16, and for PaLM 2-XXS, learning rate = 1e-5, batch size = 8.
On the TLDR summarization dataset, the final selected hyper-parameters for Gemini-Nano-1 is learning rate = 1e-4, batch size = 32.
For Gemini-Nano-2, learning rate = 3e-6, batch size = 16, and for PaLM 2-XXS, learning rate = 3e-4, batch size = 8.
We train all these backbone RMs on in-house infrastructure with TPU v5. 

For the open-source LLM model (Gemma 2B), we set the learning rate to 3e-6 for QAFeedback and UltraFeedback and 5e-6 for TL;DR dataset with batch size 64, and set the training step to 3 epochs. 
We use AdamW~\citep{adamw} as the optimizer with $\beta_1=0.9$ and $\beta_2=0.98$ cosine scheduler and warmup for the first 3\% steps. 
All the experiments are done using on 4 NVIDIA A100 40G GPUs.

\section{Experiments on Mistral-7B}\label{app:exp_mistral}
To further demonstrate our method's generalization ability to the open-source model, we add one experiment with the Mistral-7B model. 
Following the same experiment protocol, we first do SFT on Mistral-7B-v0.1 to for RM initialization and then fine the RM using synthetically generated data.
On the Ultra-feedback dataset, we observe a performance improvement from 87.43 to 87.9 and on the TLDR\_summarization dataset, we improve the performance from 74.65 to 75.01. 
Both improvements are statistically significant (under paired samples $t$-test with $p$-value < 0.05).

\section{Fine-grained Performance Improvement}\label{app:fg_improvements}
As \ours generates preference data along a set of evaluation aspects, we conduct one further experiment to check how these syntactic data improve the RM training in a more fine-grained way.
Our experiments on the Ultra Feedback dataset, which provides per-aspect human ratings, show \ours most significantly improves the ``Honesty'' aspect (+1.67 rating), while the ``Verbalized Calibration'' aspect gets the least improvement (+0.12 rating). 
This observation also partially demonstrates the quality of \ours generated synthetic data, because otherwise, we will witness some performance drops on at least one aspect.

\section{Cost of RMBoost}\label{app:cost_of_rmboost}
Using OpenAI's bulk inference APIs, the QA-feedback dataset costs \$1.638 for GPT-3.5 and \$41.417 for GPT-4; the TLDR summarization dataset costs \$5.71 for GPT-3.5 and \$148.79 for GPT-4. 
These numbers are calculated based the OpenAI's pricing strategy back around April 2024.
The costs for using PaLM 2-L (from public APIs) are on a similar scale to GPT-4. We believe these methods are generally more cost-effective than expert curations.

\section{Scaling Law of Synthetic Data}\label{app:syn_scaling_law}
When designing our data mixture strategy, we initially aimed for a 1:1 ratio of synthetic to real data. In practice, however, we apply automated rules to filter out low-quality synthetic pairs. As a result, in the ``Real+Syn'' configuration, synthetic data typically comprises between 30-50\% of the total dataset. A pilot study using the QA feedback dataset, where synthetic and real data were mixed at a 7:1 ratio, revealed that such an imbalance can actually degrade performance.

We also observe that training reward models solely on synthetic data generally underperforms compared to training on real data alone. 
Two factors may explain this. 
First, our evaluation is based on human-rated test datasets, and the real training data, which is also human-rated, is more likely to share a similar distribution with the test data. Second, the results presented in the main text were derived from reward models trained with an approximately equal mix of synthetic and real data. 
Although a pilot study using only synthetic data from the QA feedback dataset, amounting to roughly three times the volume of real data, showed some performance improvements, the process involved injecting significant human insights through ad hoc filtering rules. 
This introduces potential biases, so we opted not to include these results in our main experiments to ensure fairness.

\section{Data Generation Prompts}\label{app:prompts}

Below we list the prompts used for all compared methods across three datasets. 
Specifically, for each dataset, we have one prompt used in RLAIF (for labeling two responses side-by-side), two prompts used in RLCD (one for generating the preferred response and the other for generating the less preferred response), and two prompts used in \ours (one for generating the first response and the other for preference-conditional generation of the second response).
For \ours, we show the case where the first generated response is more preferred while the second one is less preferred. By switching the calling order of these two prompts (plus a few small word changes), we can obtain the case where the first response is less preferred than the second one.

\subsection{QA Feedback}\label{app_subsec:qa_feedback}

\subsubsection{RLAIF}\label{app:qa_feedback_RLAIF_prompt}

\begin{lstlisting}[style=mystyle, caption={Instruction used by RLAIF for QA Feedback Dataset}, label=lst:qa_feedback_rlaif_prompt, escapeinside={<@}{@>}]
You work as a professor in a top-tier university. Your goal is to teach students how to respond to a complex question given a set of Wikipedia passages as context. In this task, you will be presented with a question, a set of Wikipedia passages, a reference response, and two candidate responses that suppose to answer the given question. Your goal is to compare these two candidate responses from a set of evaluation aspects and decide which one is better for each evaluation aspect.

<task_description>
Below you will first see a guideline with detailed evaluation aspects of the response. Then, you are presented with the question, the set of Wikipedia passages, a reference response, and two candidate responses. After that, for each aspect, please judge if one candidate response is better than the other. Finally, you need to give an overall recommendation on which candidate response is better. Think about your answers first before making the judgement.
</task_description>

<guideline>
A good response to the question should provide both answer(s) that directly responds to the qeustion and crucial auxiliary information for better comprehension of the answer(s). We consider auxiliary information as crucial if it is used in the reference response. Additionally, all information in a good response should be factually consistent with (i.e., grounded in) the passages. Note that the reference response is written by a human with potentially different grounding passages, and thus, you might find answers that can be found in the passages but are not included in the reference, which is considered as acceptable. On the other hand, answers in the reference that cannot be found in or verifiable by the passages are NOT expected to be in a good response. To conclude, all answers are expected in a good response IF AND ONLY IF it can be found in the passages. Crucial auxiliary information is expected in a good response IF AND ONLY IF it can be found in both the reference response and the passages.

We will evaluate a response from the following aspects:
- (Relevance and Coherence): Whether the response contains irrelevant information (e.g., neither an answer nor crucial auxiliary information) and whether the response contains major grammar error (ignore minor typos), is uninterpretable, contradicts to common sense, or is not coherent with its context.
- (Factuality and Faithfulness): Whether the response is factually consistent with the passages and contains information that is factually verifiable. Common sense (e.g., "a bicycle has two wheels") doesn't need to be verified. However, do not count knowledge only commonly known in a specific region/community as common sense.
- (Completeness): Whether the response contains all needed information including both the answer(s) that directly responds to the qeustion and crucial auxiliary information mentioned in the reference response.
</guideline>

Below is the question.
<question>
<@\textcolor{blue}{[Question]}@>
</question>

Below is the Wikipedia passages as the context for answering the above question.
<passages>
<@\textcolor{blue}{[Passage]}@>
</passages>

Below is the reference response.
<reference_response>
<@\textcolor{blue}{[Reference Response]}@>
</reference_response>

Below is the first candidate response.
<first_response>
<@\textcolor{blue}{[First Response]}@>
</first_response>

Below is the second candidate response.
<second_response>
<@\textcolor{blue}{[Second Response]}@>
</second_response>

Read the question, passages, reference response, and two candidate responses carefully. Review the above task description and guideline. Please briefly recite your tasks back to me in your own words. For each aspect, first decide if one candidate response is better than the other and then write a one sentence explanation on why the selected response is better than the other one or why two responses are about the same. For each aspect, put the decision in <ASPECT_NAME_answer></ASPECT_NAME_answer> tags and the explanation in <ASPECT_NAME_explanation></ASPECT_NAME_explanation> tags. Note here the ASPECT_NAME in the tags should be replaced with the real aspect name (i.e. one of ["Relevance and Coherence", "Factuality and Faithfulness", "Completeness"]). Finally, make an overall decision on which candidate response is better. Put the overall decision in <final_answer></final_answer> tags and the explanation in <final_explanation></final_explanation> tags. All decisions should be in ["first_response", "second_response", "the_same"].
\end{lstlisting}

\subsubsection{RLCD}\label{app:qa_feedback_rlcd_prompt}

\begin{lstlisting}[style=mystyle, caption={Instruction used by RLCD for generating good QA Feedback response}, label=lst:qa_feedback_rlcd_positive_prompt, escapeinside={<@}{@>}]
You work as a professor in a top-tier university. You will be presented with the question, a set of Wikipedia passages, a reference response, and aim to produce another good response to the question.

<guideline>
A good response to the question should provide both answer(s) that directly responds to the qeustion and crucial auxiliary information for better comprehension of the answer(s). We consider auxiliary information as crucial if it is used in the reference response. Additionally, all information in a good response should be factually consistent with (i.e., grounded in) the passages. Note that the reference response is written by a human with potentially different grounding passages, and thus, you might find answers that can be found in the passages but are not included in the reference, which is considered as acceptable. On the other hand, answers in the reference that cannot be found in or verifiable by the passages are NOT expected to be in a good response. To conclude, all answers are expected in a good response IF AND ONLY IF it can be found in the passages. Crucial auxiliary information is expected in a good response IF AND ONLY IF it can be found in both the reference response and the passages.

A good response should have the following qualities:
- (Relevance and Coherence): The response should not contain irrelevant information (e.g., neither an answer nor crucial auxiliary information) and the response should not contain major grammar error (ignore minor typos). The response should be interpretable, follows common sense, and coherent with its context.
- (Factuality and Faithfulness): The response should be consistent with the passages and contains information that is factually verifiable. Common sense (e.g., "a bicycle has two wheels") doesn't need to be verified. However, the response should not count knowledge only commonly known in a specific region/community as common sense.
- (Completeness): The response should contain all needed information including both the answer(s) that directly responds to the qeustion and crucial auxiliary information mentioned in the reference response.
</guideline>

Below is the question.
<question>
<@\textcolor{blue}{[Question]}@>
</question>

Below is the Wikipedia passages as the context for answering the above question.
<passages>
<@\textcolor{blue}{[Passage]}@>
</passages>

Below is the reference response.
<reference_response>
<@\textcolor{blue}{[Reference Response]}@>
</reference_response>

Read the question, passages, and reference response carefully. Review the above guideline. Think about how to accomplish the task step by step before you reply. Put your generated good response in <response></response> tags.
\end{lstlisting}

\begin{lstlisting}[style=mystyle, caption={Instruction used by RLCD for generating bad QA Feedback response}, label=lst:qa_feedback_rlcd_negative_prompt, escapeinside={<@}{@>}]
You work as a professor in a top-tier university. You will be presented with the question, a set of Wikipedia passages, a reference response, and aim to produce another bad response to the question.

<guideline>
A bad response to the question misses the answer(s) that directly responds to the qeustion or crucial auxiliary information for better comprehension of the answer(s). We consider auxiliary information as crucial if it is used in the reference response. Additionally, some information in the bad response may be factually inconsistent with (i.e., not grounded in) the passages. Note that the reference response is written by a human with potentially different grounding passages, and thus, you might find answers that can be found in the passages but are not included in the reference, which is considered as acceptable. On the other hand, answers in the reference that cannot be found in or verifiable by the passages are NOT expected to be in a good response. 

A bad response will have at least one of the following qualities:
- (Bad Relevance and Coherence): The response contains irrelevant information (e.g., neither an answer nor crucial auxiliary information) or contains major grammar error (ignore minor typos). The response is uninterpretable, contradicts to common sense, or is not coherent with its context.
- (Bad Factuality and Faithfulness): The response is not factually consistent with the passages and contains information that is not factually verifiable. Common sense (e.g., "a bicycle has two wheels") doesn't need to be verified. However, knowledge only commonly known in a specific region/community is not considered as common sense.
- (Bad Completeness): The response does not contain all needed information including both the answer(s) that directly responds to the qeustion and crucial auxiliary information mentioned in the reference response.
</guideline>

Below is the question.
<question>
<@\textcolor{blue}{[Question]}@>
</question>

Below is the Wikipedia passages as the context for answering the above question.
<passages>
<@\textcolor{blue}{[Passage]}@>
</passages>

Below is the reference response.
<reference_response>
<@\textcolor{blue}{[Reference Response]}@>
</reference_response>

Read the question, passages, and reference response carefully. Review the above guideline. Think about how to accomplish the task step by step before you reply. Put your generated bad response in <response></response> tags.
\end{lstlisting}

\subsubsection{RMBoost}\label{app:qa_feedback_rmboost_prompt}

\begin{lstlisting}[style=mystyle, caption={Instruction used by RMBoost for generating the first QA Feedback response}, label=lst:qa_feedback_rmboost_y1_prompt, escapeinside={<@}{@>}]
You work as a professor in a top-tier university. Your goal is to teach students how to respond to a complex question given a set of Wikipedia passages as context. You will be presented with the question, a set of Wikipedia passages, a reference response, and aim to produce another good response to the question.

<task_description>
Below you will first see a guideline with detailed evaluation aspects of the response. Then, you are presented with the question, the set of Wikipedia passages, and a reference response. After that, you need to generate a good response to the question. This good response should NOT directly copy the reference response but is roughly of the same length as the reference response.
</task_description>

<guideline>
A good response to the question should provide both answer(s) that directly responds to the qeustion and crucial auxiliary information for better comprehension of the answer(s). We consider auxiliary information as crucial if it is used in the reference response. Additionally, all information in a good response should be factually consistent with (i.e., grounded in) the passages. Note that the reference response is written by a human with potentially different grounding passages, and thus, you might find answers that can be found in the passages but are not included in the reference, which is considered as acceptable. On the other hand, answers in the reference that cannot be found in or verifiable by the passages are NOT expected to be in a good response. To conclude, all answers are expected in a good response IF AND ONLY IF it can be found in the passages. Crucial auxiliary information is expected in a good response IF AND ONLY IF it can be found in both the reference response and the passages.

We will evaluate a response from the following aspects:
- (Relevance and Coherence): Whether the response contains irrelevant information (e.g., neither an answer nor crucial auxiliary information) and whether the response contains major grammar error (ignore minor typos), is uninterpretable, contradicts to common sense, or is not coherent with its context.
- (Factuality and Faithfulness): Whether the response is factually consistent with the passages and contains information that is factually verifiable. Common sense (e.g., "a bicycle has two wheels") doesn't need to be verified. However, do not count knowledge only commonly known in a specific region/community as common sense.
- (Completeness): Whether the response contains all needed information including both the answer(s) that directly responds to the qeustion and crucial auxiliary information mentioned in the reference response.
</guideline>

Below is the question.
<@\textcolor{blue}{[Question]}@>
{input_question}
</question>

Below is the Wikipedia passages as the context for answering the above question.
<passages>
<@\textcolor{blue}{[Passage]}@>
</passages>

Below is the reference response.
<reference_response>
<@\textcolor{blue}{[Reference Response]}@>
</reference_response>

Read the question, passages, and reference response carefully. Review the above task description and guideline. Think about how to accomplish the task step by step before you reply. Put your generated response in <response></response> tags.
\end{lstlisting}

\begin{lstlisting}[style=mystyle, caption={Instruction used by RMBoost for generating the second (less preferred) QA Feedback response}, label=lst:qa_feedback_rmboost_y2_prompt, escapeinside={<@}{@>}]
You work as a professor in a top-tier university. Your goal is to teach students how to answer a complex question given a set of Wikipedia passages. You will be presented with the question, a set of Wikipedia passages, a reference response, a good response, and aim to produce another response worse than the given good response for some aspects.

<task_description>
Below you will first see a guideline with detailed evaluation aspects of the response. Then, you are presented with the question, the set of Wikipedia passages, a reference response, and a good response. After that, you need to accomplish the following steps:
Step 1: Select a few aspects in the guideline.
Step 2: Generate another response that is worse than the good response in terms of above selected aspects.
</task_description>

<guideline>
A good response to the question should provide both answer(s) that directly responds to the qeustion and crucial auxiliary information for better comprehension of the answer(s). We consider auxiliary information as crucial if it is used in the reference response. Additionally, all information in a good response should be factually consistent with (i.e., grounded in) the passages. Note that the reference response is written by a human with potentially different grounding passages, and thus, you might find answers that can be found in the passages but are not included in the reference, which is considered as acceptable. On the other hand, answers in the reference that cannot be found in or verifiable by the passages are NOT expected to be in a good response. To conclude, all answers are expected in a good response IF AND ONLY IF it can be found in the passages. Crucial auxiliary information is expected in a good response IF AND ONLY IF it can be found in both the reference response and the passages.

We will evaluate a response from the following aspects:
- (Relevance and Coherence): Whether the response contains irrelevant information (e.g., neither an answer nor crucial auxiliary information) and whether the response contains major grammar error (ignore minor typos), is uninterpretable, contradicts to common sense, or is not coherent with its context.
- (Factuality and Faithfulness): Whether the response is factually consistent with the passages and contains information that is factually verifiable. Common sense (e.g., "a bicycle has two wheels") doesn't need to be verified. However, do not count knowledge only commonly known in a specific region/community as common sense.
- (Completeness): Whether the response contains all needed information including both the answer(s) that directly responds to the qeustion and crucial auxiliary information mentioned in the reference response.
</guideline>

Below is the question.
<question>
<@\textcolor{blue}{[Question]}@>
</question>

Below is the Wikipedia passages as the context for answering the above question.
<passages>
<@\textcolor{blue}{[Passage]}@>
</passages>

Below is the reference response.
<reference_response>
<@\textcolor{blue}{[Reference Response]}@>
</reference_response>

Below is the good response.
<good_response>
<@\textcolor{green}{[First Generated Good Response]}@>
</good_response>

Read the question, passages, reference response, and the good response carefully. Review the above task description and guideline. Think about how to accomplish the task step by step before you reply. Put your generated bad response in <response></response> tags.
\end{lstlisting}

\subsection{Ultra Feedback}\label{app_subsec:ultra_feedback}

\subsubsection{RLAIF}\label{app:ultra_feedback_RLAIF_prompt}

\begin{lstlisting}[style=mystyle, caption={Instruction used by RLAIF for Ultra Feedback Dataset}, label=lst:ultra_feedback_rlaif_prompt, escapeinside={<@}{@>}]
You are a large language model researcher. Your goal is to train a language model that follows the user input instruction with different system prompts. In this task, you will be presented with a user input instruction, a system prompt, and two candidate responses that suppose to follow the user input instruction. Your goal is to compare these two candidate responses from a set of evaluation aspects and decide which one is better for each evaluation aspect.

<task_description>
Below you will first see a guideline with detailed evaluation aspects of the response. Then, you are presented with the instruction, the system prompt, and two candidate responses. After that, for each aspect, please judge if one candidate response is better than the other. Finally, you need to give an overall recommendation on which candidate response is better. Think about your answers first before making the judgement.
</task_description>

<guideline>
We will evaluate a response from the following aspects:
- (Honesty): The assistant should be honest about whether it knows the answer and express its uncertainty explicitly. Be confident on questions it knows well and be modest on those it is unfamiliar with. Use weakeners such as 'I guess', 'I suppose', 'probably', and 'perhaps' to express uncertainty, and feel free to answer 'I don’t know' if necessary.
- (Truthfulness): The assistant should answer truthfully and be faithful to factual knowledge as well as given contexts, never making up any new facts that aren’t true or cannot be grounded in the instruction.
- (Faithful to input): The article should be faithful to the original press release without adding unsupported information or inaccurate statements.
- (Helpfulness): The assistant should provide users with accurate, relevant, and up-to-date information, ensuring that the content is positive, interesting, engaging, educational, and helpful.
- (Verbalized Calibration): The assistant should express its confidence as a scalar at the end of the response. The confidence level indicates the degree of certainty it has about its answer and is represented as a percentage.
</guideline>

Below is the system prompt.
<system_prompt>
<@\textcolor{blue}{[System Prompt]}@>
</system_prompt>

Below is the user input instruction.
<instruction>
<@\textcolor{blue}{[Instruction]}@>
</instruction>

Below is the first candidate response.
<first_response>
<@\textcolor{blue}{[First Response]}@>
</first_response>

Below is the second candidate response.
<second_response>
<@\textcolor{blue}{[Second Response]}@>
</second_response>

Read the user input instruction, system prompt, and two candidate responses carefully. Review the above task description and guideline. Please briefly recite your tasks back to me in your own words. For each aspect, first decide if one candidate response is better than the other and then write a one sentence explanation on why the selected response is better than the other one or why two responses are about the same. For each aspect, put the decision in <ASPECT_NAME_answer></ASPECT_NAME_answer> tags and the explanation in <ASPECT_NAME_explanation></ASPECT_NAME_explanation> tags. Note here the ASPECT_NAME in the tags should be replaced with the real aspect name (i.e. one of ["Honesty", "Truthfulness", "Faithful to input", "Helpfulness", "Verbalized Calibration"]). Finally, make an overall decision on which candidate response is better. Put the overall decision in <final_answer></final_answer> tags and the explanation in <final_explanation></final_explanation> tags. All decisions should be in ["first_response", "second_response", "the_same"].
\end{lstlisting}

\subsubsection{RLCD}\label{app:ultra_feedback_RLCD_prompt}

\begin{lstlisting}[style=mystyle, caption={Instruction used by RLCD for generating good Ultra Feedback response}, label=lst:ultra_feedback_rlcd_positive_prompt, escapeinside={<@}{@>}]
You are a large language model researcher. Your goal is to train a good model that follows the user input instruction with different system prompts. You will be presented with the instruction, the system prompt, and aim to produce one good response.

<guideline>
A good response should come from an assistant that has the following qualities:
- (Honesty): The assistant should be honest about whether it knows the answer and express its uncertainty explicitly. Be confident on questions it knows well and be modest on those it is unfamiliar with. Use weakeners such as 'I guess', 'I suppose', 'probably', and 'perhaps' to express uncertainty, and feel free to answer 'I don’t know' if necessary.
- (Truthfulness): The assistant should answer truthfully and be faithful to factual knowledge as well as given contexts, never making up any new facts that aren’t true or cannot be grounded in the instruction.
- (Faithful to input): The article should be faithful to the original press release without adding unsupported information or inaccurate statements.
- (Helpfulness): The assistant should provide users with accurate, relevant, and up-to-date information, ensuring that the content is positive, interesting, engaging, educational, and helpful.
- (Verbalized Calibration): The assistant should express its confidence as a scalar at the end of the response. The confidence level indicates the degree of certainty it has about its answer and is represented as a percentage.
</guideline>

Below is the system prompt.
<system_prompt>
<@\textcolor{blue}{[System Prompt]}@>
</system_prompt>

Below is the user input instruction.
<instruction>
<@\textcolor{blue}{[Instruction]}@>
</instruction>

Read the system prompt and instruction carefully. Review the above guideline. Think about your output response first before you respond. Put your generated good response in <response></response> tags.
\end{lstlisting}

\begin{lstlisting}[style=mystyle, caption={Instruction used by RLCD for generating bad Ultra Feedback response}, label=lst:ultra_feedback_rlcd_negative_prompt, escapeinside={<@}{@>}]
You are a large language model researcher. Your goal is to train a good model that follows the user input instruction with different system prompts. You will be presented with the instruction, the system prompt, and aim to produce one bad response that a good LLM will not generate.

<guideline>
A bad response should come from an assistant that has at least one of the following qualities:
- (Bad Honesty): The assistant is not honest about whether it knows the answer and fails to express its uncertainty explicitly. The assistant is over-confident on questions it doesn't know well and is not modest on those it is unfamiliar with. 
- (Bad Truthfulness): The assistant doesn't answer truthfully and is not faithful to factual knowledge as well as given contexts. The assistant makes up some new facts that aren’t true or cannot be grounded in the instruction.
- (Not Faithful to input): The assistant produces response that is not faithful to the original press release or adds unsupported information and inaccurate statements.
- (Not Helpful): The assistant doesn't provide users with accurate, relevant, and up-to-date information. The assistant fails to output content that is positive, interesting, engaging, educational, and helpful.
- (Not Verbalized Calibration): The assistant fails to express its confidence as a scalar at the end of the response. The confidence level indicates the degree of certainty it has about its answer and is represented as a percentage.
</guideline>

Below is the system prompt.
<system_prompt>
<@\textcolor{blue}{[System Prompt]}@>
</system_prompt>

Below is the user input instruction.
<instruction>
<@\textcolor{blue}{[Instruction]}@>
</instruction>

Read the system prompt and instruction carefully. Review the above guideline. Think about your output response first before you respond. Put your generated bad response in <response></response> tags.
\end{lstlisting}

\subsubsection{RMBoost}\label{app:ultra_feedback_RMBoost_prompt}

\begin{lstlisting}[style=mystyle, caption={Instruction used by RMBoost for generating the first Ultra Feedback response}, label=lst:ultra_feedback_rmboost_y1_prompt, escapeinside={<@}{@>}]
You are a large language model researcher. Your goal is to train a good model that follows the user input instruction with different system prompts. You will be presented with the instruction, the system prompt, and aim to produce one good response.

<task_description>
Below you will first see a guideline with detailed evaluation aspects of the response. Then, you are presented with the system prompt, the instruction. After that, you need to generate a good response to the question.
</task_description>

<guideline>
A response is evaluated from the following aspects:
- (Honesty): The assistant should be honest about whether it knows the answer and express its uncertainty explicitly. Be confident on questions it knows well and be modest on those it is unfamiliar with. Use weakeners such as 'I guess', 'I suppose', 'probably', and 'perhaps' to express uncertainty, and feel free to answer 'I don’t know' if necessary.
- (Truthfulness): The assistant should answer truthfully and be faithful to factual knowledge as well as given contexts, never making up any new facts that aren’t true or cannot be grounded in the instruction.
- (Faithful to input): The article should be faithful to the original press release without adding unsupported information or inaccurate statements.
- (Helpfulness): The assistant should provide users with accurate, relevant, and up-to-date information, ensuring that the content is positive, interesting, engaging, educational, and helpful.
- (Verbalized Calibration): The assistant should express its confidence as a scalar at the end of the response. The confidence level indicates the degree of certainty it has about its answer and is represented as a percentage.
</guideline>

Below is the system prompt.
<system_prompt>
<@\textcolor{blue}{[System Prompt]}@>
</system_prompt>

Below is the user input instruction.
<instruction>
<@\textcolor{blue}{[Instruction]}@>
</instruction>

Read the system prompt and instruction carefully. Review the above task description and guideline. Think about your output response first before you respond. Put your generated response in <response></response> tags.
\end{lstlisting}

\begin{lstlisting}[style=mystyle, caption={Instruction used by RMBoost for generating the second (less preferred) Ultra Feedback response}, label=lst:ultra_feedback_rmboost_y2_prompt, escapeinside={<@}{@>}]
You are a large language model researcher. Your goal is to train a good model that follows the user input instruction with different system prompts. You will be presented with the instruction, the system prompt, a good response and aim to produce another response worse than the given good response for some aspects.

<task_description>
Below you will first see a guideline with detailed evaluation aspects of the response. Then, you are presented with the system prompt, the instruction, and a good response. After that, you need to accomplish the following steps:
Step 1: Select a few aspects in the guideline.
Step 2: Generate another response that is worse than the good response in terms of above selected aspects.
</task_description>

<guideline>
A response is evaluated from the following aspects:
- (Honesty): The assistant should be honest about whether it knows the answer and express its uncertainty explicitly. Be confident on questions it knows well and be modest on those it is unfamiliar with. Use weakeners such as 'I guess', 'I suppose', 'probably', and 'perhaps' to express uncertainty, and feel free to answer 'I don’t know' if necessary.
- (Truthfulness): The assistant should answer truthfully and be faithful to factual knowledge as well as given contexts, never making up any new facts that aren’t true or cannot be grounded in the instruction.
- (Faithful to input): The article should be faithful to the original press release without adding unsupported information or inaccurate statements.
- (Helpfulness): The assistant should provide users with accurate, relevant, and up-to-date information, ensuring that the content is positive, interesting, engaging, educational, and helpful.
- (Verbalized Calibration): The assistant should express its confidence as a scalar at the end of the response. The confidence level indicates the degree of certainty it has about its answer and is represented as a percentage.
</guideline>

Below is the system prompt.
<system_prompt>
<@\textcolor{blue}{[System Prompt]}@>
</system_prompt>

Below is the user input instruction.
<instruction>
<@\textcolor{blue}{[Instruction]}@>
</instruction>

Below is the good response.
<good_response>
<@\textcolor{green}{[First Generated Good Response]}@>
</good_response>

Read the system prompt and instruction carefully. Review the above task description and guideline. Think about how to accomplish the task step by step before you respond. Put your generated bad response in <response></response> tags.
\end{lstlisting}

\subsection{TLDR Summarization}\label{app_subsec:tldr_summarization}

\subsubsection{RLAIF}\label{app:tldr_summarization_RLAIF_prompt}

\begin{lstlisting}[style=mystyle, caption={Instruction used by RLAIF for TLDR Summarization Dataset}, label=lst:tldr_summarization_rlaif_prompt, escapeinside={<@}{@>}]
You work as a professor in a top-tier university. You aim to teach students how to summarize a reddit post. In this task, you will be presented with a post and two candidate summaries that suppose to summarize the given post. Your goal is to compare these two candidate summaries from a set of evaluation aspects and decide which one is better for each evaluation aspect.

<task_description>
Below you will first see a guideline with detailed evaluation aspects of the summary. Then, you are presented with the post and two candidate summaries. After that, for each aspect, please judge if one candidate summary is better than the other. Finally, you need to give an overall recommendation on which candidate summary is better. Think about your answers first before making the judgement.
</task_description>

<guideline>
A good summary is a shorter piece of text that accomplishes the same purpose and conveys the same information as the original post.

We will evaluate a summary from the following aspects:
- (Coherence): A summary is coherent if, when read by itself it's easy to understand and free of English errors. A summary is not coherent if it is difficult to understand what the summary is trying to say. It's more important that the summary is understandable than it being free of grammar errors.
- (Accuracy): A summary is accurate if it doesn't say things that aren't in the post, it doesn't mix up people, and generally is not misleading. If the summary says anything that is not mentioned in the post or contradicts something in the post, this summary is not accurate enough.
- (Coverage): A summary has good coverage if it mentions the main information from the post that's important to understand the situation described in the post. A summary has poor coverage if someone reading only the summary would be missing several important pieces of information about the situation in the post.
</guideline>

Below is the post.
<post>
<@\textcolor{blue}{[Post]}@>
</post>

Below is the first candidate summary.
<first_summary>
<@\textcolor{blue}{[First Summary]}@>
</first_summary>

Below is the second candidate summary.
<second_summary>
<@\textcolor{blue}{[Second Summary]}@>
</second_summary>

Read the post and two candidate summaries carefully. Review the above task description and guideline. Please briefly recite your tasks back to me in your own words. For each aspect, first decide if one candidate summary is better than the other and then write a one sentence explanation on why the selected summary is better than the other one or why two summaries are about the same. For each aspect, put the decision in <ASPECT_NAME_answer></ASPECT_NAME_answer> tags and the explanation in <ASPECT_NAME_explanation></ASPECT_NAME_explanation> tags. Note here the ASPECT_NAME in the tags should be replaced with the real aspect name (i.e. one of ["Coherence", "Accuracy", "Coverage"]). Finally, make an overall decision on which candidate summary is better. Put the overall decision in <final_answer></final_answer> tags and the explanation in <final_explanation></final_explanation> tags. All decisions should be in ["first_summary", "second_summary", "the_same"].
\end{lstlisting}

\subsubsection{RLCD}\label{app:tldr_summarization_RLCD_prompt}

\begin{lstlisting}[style=mystyle, caption={Instruction used by RLCD for generating good TLDR Summarization response}, label=lst:tldr_summarization_rlcd_positive_prompt, escapeinside={<@}{@>}]
You work as a professor in a top-tier university. Your goal is to teach students how to summarize a reddit post. A good summary is a shorter piece of text that accomplishes the same purpose and conveys the same information as the original post. You will be presented with the post and aim to produce a good response as the summary.

<guideline>
A good summary should have the following qualities:
- (Coherence): A good summary is coherent if, when read by itself it's easy to understand and free of English errors. It's more important that the good summary is understandable than it being free of grammar errors.
- (Accuracy): A good summary is accurate if it doesn't say things that aren't in the post, it doesn't mix up people, and generally is not misleading. If the summary says anything that is not mentioned in the post or contradicts something in the post, this summary is not accurate enough.
- (Coverage): A good summary has good coverage if it mentions the main information from the post that's important to understand the situation described in the post. A summary has poor coverage if someone reading only the summary would be missing several important pieces of information about the situation in the post.
</guideline>

Below is the post.
<post>
<@\textcolor{blue}{[Post]}@>
<post>

Read the post and guideline carefully. Think about your output summary first before you respond. Put your generated good response in <response></response> tags.
\end{lstlisting}

\begin{lstlisting}[style=mystyle, caption={Instruction used by RLCD for generating bad TLDR Summarization response}, label=lst:tldr_summarization_rlcd_negative_prompt, escapeinside={<@}{@>}]
You work as a professor in a top-tier university. Your goal is to teach students how to summarize a reddit post. A good summary is a shorter piece of text that accomplishes the same purpose and conveys the same information as the original post. The bad summary fails to accomplish this goal and thus you will teach student not to produce it. Below, you will be presented with the post and aim to produce a bad response as the summary.

<guideline>
A bad summary will have at least one of the following qualities:
- (Bad Coherence): A bad summary is not coherent when read by itself it's hard to understand and has some English errors. A bad summary is not coherent as it is difficult to understand what the summary is trying to say.
- (Bad Accuracy): A bad summary is not accurate as it says something that is not mentioned in the post or contradicts something in the post.
- (Bda Coverage): A bad summary has poor coverage when someone reading only the summary would be missing several important pieces of information about the situation in the post.
</guideline>

Below is the post.
<post>
<@\textcolor{blue}{[Post]}@>
<post>

Read the post and guideline carefully. Think about your output summary first before you respond. Put your generated bad response in <response></response> tags.
\end{lstlisting}

\subsubsection{RMBoost}\label{app:tldr_summarization_RMBoost_prompt}

\begin{lstlisting}[style=mystyle, caption={Instruction used by RMBoost for generating the first TLDR Summarization response}, label=lst:tldr_summarization_rmboost_y1_prompt, escapeinside={<@}{@>}]
You work as a professor in a top-tier university. Your goal is to teach students how to summarize a reddit post. A good summary is a shorter piece of text that accomplishes the same purpose and conveys the same information as the original post. You will be presented with the post and aim to produce a good response as the summary.

<task_description>
Below you will first see a guideline with detailed evaluation aspects of the summary. Then, you are presented with the post. After that, you need to generate a good summary of the post as the response.
</task_description>

<guideline>
A summary is evaluated from the following aspects:
- (Coherence): A summary is coherent if, when read by itself it's easy to understand and free of English errors. A summary is not coherent if it is difficult to understand what the summary is trying to say. It's more important that the summary is understandable than it being free of grammar errors.
- (Accuracy): A summary is accurate if it doesn't say things that aren't in the post, it doesn't mix up people, and generally is not misleading. If the summary says anything that is not mentioned in the post or contradicts something in the post, this summary is not accurate enough.
- (Coverage): A summary has good coverage if it mentions the main information from the post that's important to understand the situation described in the post. A summary has poor coverage if someone reading only the summary would be missing several important pieces of information about the situation in the post.
</guideline>

Below is the post.
<post>
<@\textcolor{blue}{[Post]}@>
<post>

Read the post carefully. Review the above task description and guideline. Think about your output summary first before you respond. Put your generated response in <response></response> tags.
\end{lstlisting}

\begin{lstlisting}[style=mystyle, caption={Instruction used by RMBoost for generating the second (less preferred) TLDR Summarization response}, label=lst:tldr_summarization_rmboost_y2_prompt, escapeinside={<@}{@>}]
You work as a professor in a top-tier university. Your goal is to teach students how to summarize a reddit post. A good summary is a shorter piece of text that accomplishes the same purpose and conveys the same information as the original post. You will be presented with the post, a good response, and aim to produce another response worse than the given good response for some aspects.

<task_description>
Below you will first see a guideline with detailed evaluation aspects of the summary. Then, you are presented with the post and a good response. After that, you need to accomplish the following steps:
Step 1: Select a few aspects in the guideline.
Step 2: Generate another response that is worse than the first response in terms of above selected aspects. 
</task_description>

<guideline>
A summary is evaluated from the following aspects:
- (Coherence): A summary is coherent if, when read by itself it's easy to understand and free of English errors. A summary is not coherent if it is difficult to understand what the summary is trying to say. It's more important that the summary is understandable than it being free of grammar errors.
- (Accuracy): A summary is accurate if it doesn't say things that aren't in the post, it doesn't mix up people, and generally is not misleading. If the summary says anything that is not mentioned in the post or contradicts something in the post, this summary is not accurate enough.
- (Coverage): A summary has good coverage if it mentions the main information from the post that's important to understand the situation described in the post. A summary has poor coverage if someone reading only the summary would be missing several important pieces of information about the situation in the post.
</guideline>

Below is the post.
<post>
<@\textcolor{blue}{[Post]}@>
<post>

Below is the good response.
<good_response>
<@\textcolor{green}{[Generated First Good Response]}@>
<good_response>

Read the post carefully. Review the above task description and guideline. Think about how to accomplish the task step by step before you reply. Put your generated bad summary in <response></response> tags.
\end{lstlisting}